\documentclass[final]{cvpr}

\usepackage{times}
\usepackage{epsfig}
\usepackage{graphicx}
\usepackage{subfigure}
\usepackage{float}
\usepackage{amsmath}
\usepackage{amssymb}
\usepackage{booktabs}
\usepackage{multirow}
\usepackage{bbding}
\usepackage{pifont}
\usepackage{mathrsfs}

\usepackage[pagebackref=true,breaklinks=true,colorlinks,bookmarks=false]{hyperref}

\def\cvprPaperID{****} 
\def\confYear{CVPR 2021}

\begin{document}

\title{RGB Stream Is Enough for Temporal Action Detection}

\author{
 Chenhao Wang\thanks{Equal contribution.}, Hongxiang Cai\footnotemark[1], Yuxin Zou\thanks{Data analysis.}, Yichao Xiong\thanks{Corresponding author.} \\
  Media Intelligence Technology Co.,Ltd\\
  \texttt{\small \{chenhao.wang, hongxiang.cai, yuxin.zou, yichao.xiong\}@media-smart.cn} 
}

\maketitle

\begin{abstract}
State-of-the-art temporal action detectors to date are based on two-stream input including RGB frames and optical flow. Although combining RGB frames and optical flow boosts performance significantly, optical flow is a hand-designed representation which not only requires heavy computation, but also makes it methodologically unsatisfactory that two-stream methods are often not learned end-to-end jointly with the flow. In this paper, we argue that optical flow is dispensable in high-accuracy temporal action detection and image level data augmentation (ILDA) is the key solution to avoid performance degradation when optical flow is removed. To evaluate the effectiveness of ILDA, we design a simple yet efficient one-stage temporal action detector based on single RGB stream named DaoTAD. Our results show that when trained with ILDA, DaoTAD has comparable accuracy with all existing state-of-the-art two-stream detectors while surpassing the inference speed of previous methods by a large margin and the inference speed is astounding 6668 fps on GeForce GTX 1080 Ti. Code is available at \url{https://github.com/Media-Smart/vedatad}.
\end{abstract}

\section{Introduction}
Over recent years, with rapid development of Internet, a dramatically growing number of videos have been generated and stored in online platforms, which makes video content analysis attract much interest from both academia and industry. As a fundamental task in video analysis field, action recognition aims to classify well-trimmed video clips and has made great progress due to the vast success of deep learning. Compared with action recognition, temporal action detection (TAD) is more challenging and the goal is to detect action instances including action categories and temporal boundaries (start time, end time) in untrimmed videos. Temporal action detection can be applied in several practical applications such as video recommendation, smart surveillance, human interaction, \etc.

\begin{figure}[htp]
    \centering
    \includegraphics[width=8.3cm]{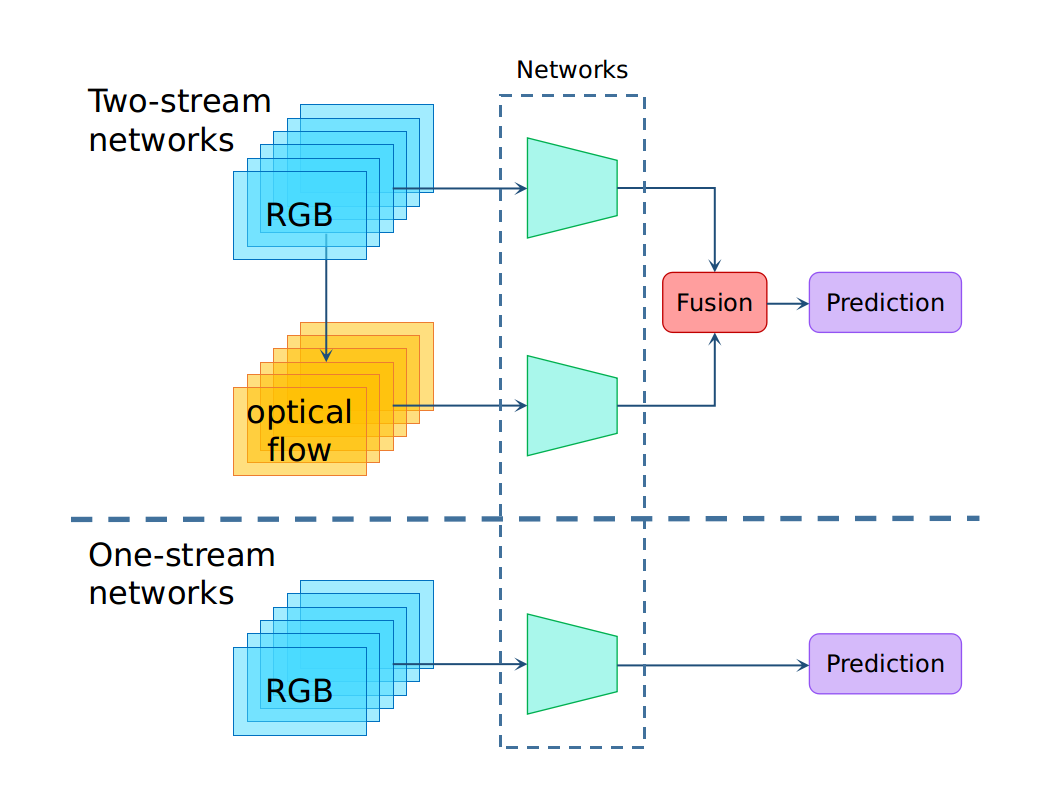}
    \caption{Illustration of one-stream and two-stream scheme. Two-stream networks are not only structurally more complicated, but also very time consuming compared to one-stream networks}
    \label{scheme compare}
\end{figure}

Similar to object detection, temporal action detection belongs to the visual detection scope. While the former aims at predicting bounding boxes to surround object instances precisely, the latter's purpose is to generate pairs of time points for segmenting action instances correctly. Therefore, most state-of-the-art methods \cite{xu2017r, lin2017single, zhang2018s3d, chao2018rethinking, wang2020multi, liu2020progressive, lin2021learning} for temporal action detection take inspirations from object detection \cite{liu2016ssd, cai2018cascade, tian2019fcos}. Generally, these methods can be divided into two-stage and one-stage. Two-stage methods \cite{xu2017r, chao2018rethinking, lin2021learning} follow the paradigm of proposal-then-classification, and first generate several segment proposals, then classify each proposal with specified video classifiers. In contrast, one-stage methods \cite{lin2017single, zhang2018s3d, liu2020progressive, wang2020multi} eliminate the time-consuming proposal generation and are generally more efficient. These methods are usually based on two-stream input including RGB frames and optical flow, and the performance degrades dramatically when optical flow is removed. For example, AFSD \cite{lin2021learning} achieves 52.0\% average mAP in [0.3:0.1:0.7] on THUMOS14 \cite{jiang2014thumos} with two-stream input while the performance drops to 43.5\% when the input is single RGB stream. MLTPN \cite{wang2020multi} has also a 8.7\% gap (48.0\% \vs 39.3\%) between two-stream input and single RGB stream input. Although combining RGB frames and optical flow has a remarkable improvement of performance, the optical flow extraction is very time-consuming which makes these methods not efficient and hard to be applied into real applications. Take Dual TV-$L^{1}$ \cite{zach2007duality} for example, extracting optical flow with shape 96$\times$96 requires 650 ms on Intel(R) Xeon(R) CPU E5-2630 v4@2.20GHz. In the meanwhile, it is methodologically unsatisfactory that two-stream methods are often not learned end-to-end jointly and may lead to sub-optimal performance.

In this paper, based on above observations, we propose a method to get rid of the dependency on optical flow without sacrificing performance. We delve into previous works and find the critical factor is image level data augmentation (ILDA), which has been neglected by the community for a long time. To verify the effectiveness of ILDA, we build a high-efficiency one-stage detector based on single RGB stream called DaoTAD, which draws on a variety of recent ideas from object detection and performs as well as two-stream methods.

In summary, our work has three main contributions as follow:
\begin{itemize}
\item We demonstrate that optical flow is dispensable in high-accuracy temporal action detection and image level data augmentation (ILDA) is the key solution to avoid performance degradation when optical flow is removed.
\item We propose a simple yet efficient one-stage temporal action detector based on single RGB stream and achieve the inference speed of 6668 fps on GeForce GTX 1080 Ti.
\item Our proposed method has comparable accuracy with all existing state-of-the-art two-stream methods and is the best one against previous one-stage methods on THUMOS14 \cite{jiang2014thumos}.
\end{itemize}

\section{Related Work}
In this section, we review the prior work relevant to action recognition, object detection and temporal action detection.

\textbf{Action Recognition}
As same as image recognition in 2D image analysis area, action recognition is a fundamental yet significant task in video understanding domain. Traditional approaches \cite{dalal2006human, klaser2008spatio, scovanner20073, wang2013action} are mostly based on hand-crafted visual features. With the vast success of deep learning methods in recent years, the majority of current approaches \cite{tran2015learning, wang2016temporal, carreira2017quo, tran2018closer, wang2018appearance, sun2018optical, wang2018non, jiang2019stm, lin2019tsm, kwon2020motionsqueeze} adopt deep neural networks and achieve superior performance. These action recognition methods assume well-trimmed video clips and are not suitable for untrimmed videos containing multiple actions. However, just as image recognition networks are used in image object detection, pre-trained action recognition models are widely used in temporal action detection for effective feature extraction. In this paper, we use the ResNet-50 I3D \cite{wang2018non} model pre-trained on Kinetics \cite{kay2017kinetics} as our basic feature extraction network.

\textbf{Object Detection}
Object detection aims to locate and classify existing objects in a 2D image. In recent years, deep learning based methods dominate this field and are mainly divided into two categories: two-stage and one-stage. Two-stage methods follow  proposal-then-classification paradigm, starting from R-CNN \cite{girshick2014rich} and its variations \cite{girshick2015fast, ren2016faster}. R-CNN \cite{girshick2014rich} uses selective search \cite{uijlings2013selective} to generate multiple region proposals and then uses CNN to classify these proposals. Faster R-CNN \cite{ren2016faster} replaces selective search with RPN and thus the whole network can be trained end-to-end. Based on Faster R-CNN \cite{ren2016faster}, extensive methods have been proposed such as FPN \cite{lin2017feature}, Mask R-CNN \cite{he2017mask}, Cascade R-CNN \cite{cai2018cascade}, IoU-Net \cite{jiang2018acquisition}, \etc. Unlike two-stage methods, one-stage methods directly detect objects without proposal generation. In the last few years, numerous approaches \cite{redmon2016you, redmon2017yolo9000, redmon2018yolov3, liu2016ssd, lin2017focal, tian2019fcos, wu2019iou, li2020generalized, wu2020iou, zheng2020distance} have been proposed in this direction to boost performance. RetinaNet \cite{lin2017focal}, a representative method among them, devises focal loss to address the extreme imbalance between foreground and background classes during training. Our network shares many similarities with RetinaNet \cite{lin2017focal}.

\begin{figure*}[htp]
    \centering
    \includegraphics[width=17.4cm]{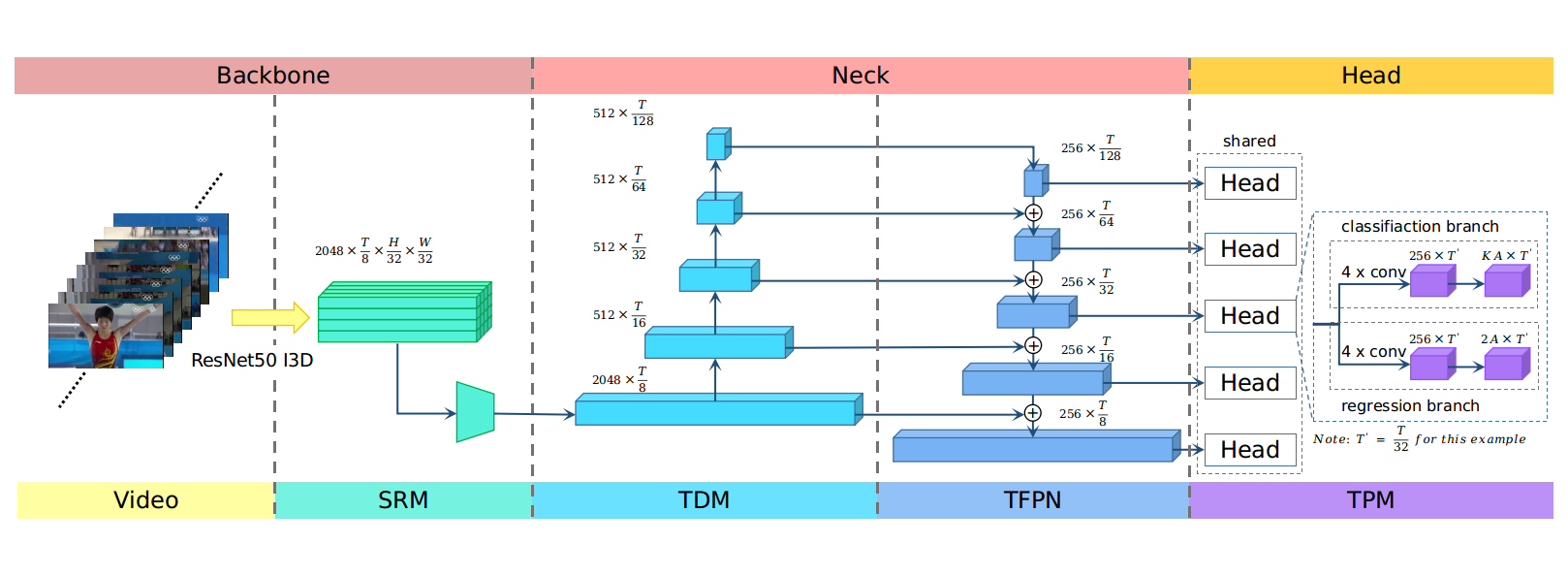}
    \caption{Overview architecture of DaoTAD. The network mainly consists of three parts, feature extractor followed by SRM as the backbone, TDM with TFPN as the neck and TPM as the head. The output $T'$ of shared head equals to the temporal size of input feature sequence in each level, respectively. For example, $T'=\frac{T}{32}$ in figure since the input size is $256\times\frac{T}{32}$.}
    \label{network structure}
\end{figure*}

\textbf{Temporal Action Detection}
Analogous to object detection, the goal of temporal action detection is to detect action instances in untrimmed videos with both action categories and temporal boundaries. In many ways, the progress of temporal action detection parallels the development of object detection. 
For two-stage methods, many approaches \cite{gao2018ctap, lin2018bsn, lin2019bmn, zeng2019graph, liu2019multi, zhao2020bottom, gao2020accurate, bai2020boundary, su2020bsn++, qing2021temporal, shou2017cdc, zhao2017temporal} follow the paradigm of \cite{girshick2014rich, girshick2015fast} and treat temporal action detection as two separated stages in sequence, i.e. first generating multiple class-agnostic segment proposals with a given video, then classifying each proposal into an action category. In this direction, most methods \cite{gao2018ctap, lin2018bsn, lin2019bmn, zeng2019graph, liu2019multi, lin2020fast, zhao2020bottom, gao2020accurate, bai2020boundary, su2020bsn++, qing2021temporal} focus on improving the quality of generated proposals in the first stage, while a few others \cite{shou2017cdc, zhao2017temporal} exploit a more accurate classifier in the second stage. However, these proposed methods don't afford end-to-end training and may lead to sub-optimal performance. Inspired by Faster R-CNN \cite{ren2016faster}, R-C3D \cite{xu2017r} uses a fully-convolutional 3D ConvNet to encode the video streams and jointly optimizes proposal generation and classification. TAL-Net \cite{chao2018rethinking} improves R-C3D \cite{xu2017r} by enforcing receptive field alignment, better exploiting the temporal context of actions for both proposal generation and action classification, and explicitly considering multi-stream feature fusion. AFSD \cite{lin2021learning} employs an anchor-free method to generate coarse proposals and proposes a saliency-based refinement module with boundary consistency learning strategy to gather more valuable boundary features for each proposal refinement. 

In contrast, one-stage methods \cite{lin2017single, zhang2018s3d, liu2020progressive, wang2020multi} perform localization and classification at the same time and thus are more efficient. SSAD \cite{lin2017single} uses 1D temporal convolutional layers to skip the proposal generation step via directly detecting action instances in untrimmed video. Inspired by SSD \cite{liu2016ssd}, S$^3$D \cite{zhang2018s3d} predicts scores for the presence of activity categories in each default span and produces temporal adjustments relative to the span location to predict the precise activity duration. PBRNet \cite{liu2020progressive} proposes three cascaded detection modules for localizing action boundary more and more precisely. MLTPN \cite{wang2020multi} improves SSAD \cite{lin2017single} by enhancing the discrimination of the features in each layer with a proposed multi-level temporal feature pyramid network. 

To achieve the best performance, most of these methods are based on two-stream input including RGB frames and optical flow, while the time-consuming extraction of optical flow makes these methods hard to be applied into real applications. In this paper, we devise an accurate and efficient one-stage method which is based on single RGB stream and achieves comparable performance against previous works.

\begin{table*}[htbp]
    \caption{Structure of Temporal Prediction Module. $T'$ is the temporal dimension size depending on which layer of TFPN the feature sequence come from, $K$ is the number of classes and $A$ is the number of anchors}
    \centering
    \begin{tabular} { c | c c c | c | c | c c c | c }
        \toprule
        \multicolumn{10}{c}{input feature sequence: $ 256 \times T' $}\\
        \midrule
        \multicolumn{5}{c|}{classification branch} & \multicolumn{5}{|c}{regression branch}\\
        \midrule
        layer & kernel & stride & channel & output size & layer & kernel & stride & channel & output size \\
        \midrule
        Conv1d-1 & 3 & 1 & 256 & $ 256 \times T' $ & Conv1d-1 & 3 & 1 & 256 & $ 256 \times T' $\\
        \midrule
        Conv1d-2 & 3 & 1 & 256 & $ 256 \times T' $ & Conv1d-1 & 3 & 1 & 256 & $ 256 \times T' $\\
        \midrule
        Conv1d-3 & 3 & 1 & 256 & $ 256 \times T' $ & Conv1d-1 & 3 & 1 & 256 & $ 256 \times T' $\\
        \midrule
        Conv1d-4 & 3 & 1 & 256 & $ 256 \times T' $ & Conv1d-1 & 3 & 1 & 256 & $ 256 \times T' $\\
        \midrule
        Conv1d-class & 3 & 1 & 256 & $ K \times A \times T' $ & Conv1d-reg & 3 & 1 & 256 & $ 2 \times A \times T' $\\
        \bottomrule
    \end{tabular}
    \label{tab:TPM}
\end{table*}

\section{Image Level Data Augmentation}
\textit{Why is data augmentation needed?} Due to the huge number of parameters in deep convolutional neural network (DCNN) and insufficient data, overfitting becomes an inevitable problem. As one of the most important techniques to avoid overfitting and improve generalization, data augmentation is an explicit form of regularization and widely used during training of DCNN. The goal of data augmentation is to artificially enlarge the training dataset from existing data by various transformations, such as translation, rotation, flipping, cropping, adding noises, \etc. 

\textit{Which kinds of data augmentation should be used  in training given specific data?} The types of data augmentation are selected elaborately depending on the types of variations in data. For instance, horizontal flipping is a kind of variation in image which is expected not to affect the result of classification, thus horizontal flipping augmentation is widely used in image classification. 

\textit{What kinds of variations exist in video data of temporal action detection?} Apparently, there are two categories of variations: temporal level and image level. Specifically, temporal level variations include temporal translation, temporal scale, \etc while image level variations consist of flipping, rotation, \etc.

Depending on temporal level variations, for example, C-TCN \cite{li2020deep} proposes Random Move to simulate the montage effect of video production. However, image level variations have been neglected for a long time by the TAD community so that most of previous works don't realize the effectiveness of image level data augmentation (ILDA) for TAD model training. We conjecture the absence of ILDA in TAD is attributed to the fact that most of current methods are based on two-stream, and optical flow extraction is time-consuming, therefore optical flow needs to be generated offline in advance which makes ILDA unpractical. Without the help of ILDA, the performance boost brought by optical flow can not be ignored so that methods based on two-stream become popular. Such contradiction forms a dilemma in the research of TAD. In this paper, we jump out of this dead cycle and propose DaoTAD, a single RGB stream detector, which makes ILDA practical. ILDA significantly increases the accuracy of DaoTAD and makes DaoTAD comparable with all existing state-of-the-art two-stream detectors.

Specifically, we not only use temporal level data augmentation but also apply four widely-used image level data augmentation such as random crop, photo distortion, random rotation, and random flip in DaoTAD.

\section{DaoTAD}
In this section, we will provide a detailed description of our proposed one-stage temporal action detector based on single RGB stream. DaoTAD mainly consists of three parts as follow: a basic 3D feature extractor equipped with a spatial reduction module (SRM) as backbone, a temporal downsample module (TDM) followed by a temporal feature pyramid network (TFPN) as neck, and a shared temporal prediction module (TPM) containing both action instance classification branch and temporal boundary regression branch as head. Given an input video, we first generate video clips with constant temporal size and reshape each clip to (C, T, H, W), where C, T, H, W represents the channel, time step, height, and width respectively. Then we use the backbone network to extract a 3D feature map which is then reduced to a 1D feature sequence by pooling the last two dimensions with the SRM. Such a sequence contains the temporal and appearance information of whole video clip. Afterwards, the feature sequence is fed into the TDM and TFPN to generate proportionally temporal sized feature sequences at multiple levels. Finally the pyramid features are further utilized to output action proposal sequence via the shared TPM, which includes an action instance classifier and temporal boundary regressor. The architecture of DaoTAD is shown in Figure \ref{network structure}.

\subsection{Feature Extractor}
For feature extraction, we use ResNet50 I3D \cite{wang2018non} as our basic 3D feature extractor due to its excellent performance in action recognition and simple architecture. Specifically, the layers of ResNet50 I3D before the last average pooling layer are employed for feature extraction. Given an input video clip, a 3D feature map with shape (2048, T/8, H/32, W/32) will be generated.

\subsection{Spatial Reduction Module}
\label{subsection:SRM}
In order to get a glimpse of the whole frame to help locate and classify temporal segments and decrease the computation cost of the network afterwards, we design SRM to squeeze the spatial dimensions of 3D feature map produced by above feature extractor. There are three different reduction modes: maximum pooling , average pooling, and convolution. In this work, we choose average pooling reduction mode based on ablation study. A 3D feature map with shape (2048, T/8, H/32, W/32) will be converted to a 1D temporal feature sequence with shape (2048, T/8) via SRM.

\begin{table}[htbp]
    \caption{Structure of Temporal Downsample Module. $T$ is the temporal dimension size.}
    \centering
    \begin{tabular} { c | c c c | c }
        \toprule
        \multicolumn{5}{c}{input feature sequence: $2048 \times \frac{T}{8}$}\\
        \midrule
        layer & kernel & stride & channel & output size\\
        \midrule
        Conv1d-1 & 3 & 2 & 512 & $ 512 \times \frac{T}{16} $ \\
        \midrule
        Conv1d-2 & 3 & 2 & 512 & $ 512 \times \frac{T}{32} $ \\
        \midrule
        Conv1d-3 & 3 & 2 & 512 & $ 512 \times \frac{T}{64} $ \\
        \midrule
        Conv1d-4 & 3 & 2 & 512 & $ 512 \times \frac{T}{128} $ \\

        \bottomrule
    \end{tabular}
    \label{tab:TDM}
\end{table}

\subsection{Temporal Downsample Module}
To cope with large temporal scale variation of action instances, we build TDM to generate proportionally temporal sized feature sequences at multiple levels. Our proposed TDM consists of 4 stacked Conv1D layers with kernel size 3 and stride 2, and the scale of temporal dimension is reduced by half after each convolution layer. Together with the input temporal feature sequence, 5-layer temporal feature pyramids are constructed with corresponding temporal feature strides of [8, 16, 32, 64, 128]. In this work, we set the output channel of each layer to 512. Figure \ref{network structure} and Table \ref{tab:TDM} shows the structure and detailed parameters of TDM.
 
\subsection{Temporal Feature Pyramid Network}
TFPN is designed to combine semantically strong low-temporal-resolution features with semantically weak high-temporal-resolution features via a top-down pathway, thus the output features contain both short and long temporal  receptive field. The design of TFPN is intuitive, when localizing and classifying an action segment we not only watch the content in the time span but also the frames before and after the time span.
The output channel of each layer is set to 256 and the number of layers and temporal strides keep the same as TDM. Hopefully, lower layers of feature sequences with high temporal resolution would be responsible for the shorter temporal action instance, while higher layers with low temporal resolution for the longer ones. 

\subsection{Temporal Prediction Module}
As a one-stage detector, DaoTAD is designed to predict action categories and temporal boundaries jointly. Thus, our proposed Temporal Prediction Module (TPM) consists of two branches: classification branch to predict the probabilities of each class for each anchor, and regression branch to predict temporal segment boundaries. The classification branch is made of 4 stacked Conv1D layers with kernel size 3 and stride 1, followed by a classification output Conv1D layer. The regression branch is almost the same as the classification branch, except for the output channel of the output layer. Furthermore, the head is shared by all the 5 layers in temporal feature pyramids. The detailed parameters of TPM are depicted in Table \ref{tab:TPM}.

\subsection{Anchor}
Similar to RetinaNet \cite{lin2017focal}, we use translation-invariant anchors in this work and the anchors have the temporal size from 16 to 256 on pyramid levels $P_3$ to $P_7$, respectively. At each level, we add anchors with 5 sizes \{$2^0$,$2^{1/5}$,$2^{2/5}$,$2^{3/5}$,$2^{4/5}$\} of the original set of default anchors for dense scale coverage. 

We use the assignment rule from RetinaNet \cite{lin2017focal} which is modified with adjusted thresholds. Specifically, anchors are assigned to ground-truth action segments using the temporal Intersection-over-Union (tIoU) threshold 0.6, and to background with the tIoU lower than 0.4. Other anchors with overlap in [0.4, 0.6) will be ignored during training.

\subsection{Loss}
\textbf{Classification Loss} We use Focal Loss \cite{lin2017focal} for classification which can alleviate class imbalance problem. 

\textbf{Regression Loss} Temporal DIoU loss is chosen for action instance temporal boundary regression, the temporal version of DIoU loss \cite{zheng2020distance} is defined as 
\begin{equation}
\mathcal{L}_{DIoU} = 1 - tIoU + \frac{\rho^{2} (c, c^{gt})}{u^{2}}
\end{equation}
in which $c$ and $c^{gt}$ stand for the central point coordinates of predicted segment and ground truth segment, $\rho(\cdot )$ stands for the operator of Euclidean distance calculation, $u$ stands for the diagonal length of the smallest enclosing segment containing the two segments. Temporal D-IoU loss takes not only tIoU, but also normalized distance between central points of two segments into consideration, which can lead to better and much faster convergence.




\section{Experiments}
\subsection{Dataset and Metric}
\textbf{Dataset}
THUMOS14 \cite{jiang2014thumos} is a commonly-used dataset in domain which contains 200 validation videos and 212 test videos with labeled temporal annotations from 20 categories. The videos in training set are trimmed and can not be used for temporal action detection task. Therefore, following the standard practice, we use the validation set for training and the test set for evaluation. 

\textbf{Evaluation Metric}
We follow the official evaluation metric of THUMOS14. In all experiments, we report the mean average precision (mAP) with tIoU thresholds 0.3, 0.4, 0.5, 0.6, 0.7.

\begin{table*}[htbp]
    \caption{Performance comparison with state-of-the-art methods on THUMOS14, measured by mAP at different tIoU thresholds, and average mAP in 0.3, 0.4, 0.5, 0.6, 0.7. \ding{51} and \ding{55} means whether the method uses optical-flow or not respectively.}
    \centering
    \begin{tabular}{ c | c | c | c c c c c }
        \toprule
        Type & Method & Flow & 0.3 & 0.4 & 0.5 & 0.6 & 0.7 \\
        \midrule
        \multirow{10}*{Two-Stage} & MGG \cite{liu2019multi} & \ding{51} & 53.9 & 46.8 & 37.4 & 29.5 & 21.3 \\
        ~ & BMN \cite{lin2019bmn} & \ding{51} & 56 & 47.4 & 38.8 & 29.7 & 20.5 \\
        ~ & GCNs \cite{zeng2019graph} & \ding{51} & 63.6 & 57.8 & 49.1 & - & - \\
        ~ & DBG \cite{lin2020fast} & \ding{51} & 57.8 & 49.4 & 39.8 & 30.2 & 21.7 \\
        ~ & BC-GNN \cite{bai2020boundary} & \ding{51} & 57.1 & 49.1 & 40.4 & 31.2 & 23.1 \\
        ~ & IntraC InterC \cite{zhao2020bottom} & \ding{51} & 53.9 & 50.7 & 45.4 & 38 & 28.5 \\
        ~ & BSN++ \cite{su2020bsn++} & \ding{51} & 59.9 & 49.5 & 41.3 & 31.9 & 22.8 \\
        ~ & TCANet \cite{qing2021temporal} & \ding{51} & 60.6 & 53.2 & 44.6 & 36.8 & 26.7 \\
        ~ & AFSD \cite{lin2021learning} & \ding{51} & \textbf{67.3} & \textbf{62.4} & \textbf{55.5} & \textbf{43.7} & \textbf{31.1} \\
        \midrule
        \multirow{8}*{One-Stage} & DBS \cite{gao2019video} & \ding{51} & 50.6 & 43.1 & 34.3 & 24.4 & 14.7 \\
        ~ & GTAN \cite{long2019gaussian} & \ding{55} & 57.8 & 47.2 & 38.8 & - & - \\
        ~ & A2Net \cite{yang2020revisiting} & \ding{51} & 58.6 & 54.1 & 45.5 & 32.5 & 17.2 \\
        ~ & PBRNet \cite{liu2020progressive} & \ding{51} & 58.5 & 54.6 & 51.3 & 41.8 & 29.5 \\
        ~ & G-TAD \cite{xu2020g} & \ding{51} & 66.4 & 60.4 & 51.6 & 37.6 & 22.9 \\
        ~ & C-TCN \cite{li2020deep} & \ding{51} & \textbf{68} & 62.3 & 52.1 & - & - \\
        ~ & MLTPN \cite{wang2020multi} & \ding{51} & 66 & \textbf{62.6} & 53.3 & 37 & 21.1 \\
        ~ & Ours & \ding{55} & 62.8 & 59.5 & \textbf{53.8} & \textbf{43.6} & \textbf{30.1} \\
        \bottomrule
    \end{tabular}
    \label{tab:compare}
\end{table*}

\begin{table}[htbp]
    \caption{Inference speed comparison with state-of-the-art methods. $\ast$ means ignoring the time cost of optical flow extraction. For example, using Dual TV-$L^{1}$ \cite{zach2007duality} to extract optical flow from a pair of $96\times96$ frames will cost 650 ms on Intel(R) Xeon(R) CPU E5-2630 v4@2.20GHz}
    \centering
    \begin{tabular}{ c | c | c }
        \toprule
        Method & GPU & FPS \\
        \midrule
        R-C3D \cite{xu2017r} & TITAN Xp & $1030^{\ast }$ \\
        PBRNet \cite{liu2020progressive} & 1080Ti & $1488^{\ast }$ \\
        AFSD \cite{lin2021learning} & 1080Ti & $3259^{\ast }$ \\
        AFSD \cite{lin2021learning} & V100 & $4057^{\ast }$ \\
        \midrule
        \textbf{Ours} & 1080Ti & \textbf{6668} \\
        \bottomrule
    \end{tabular}
    \label{tab:fps}
\end{table}

\subsection{Implementation Details}
\textbf{General Setting}
For validation and test set, we decode each video with 25 frames per second (fps) and 128$\times$128 frame resolution. Considering the fact that the time span of over 99.5\% action instances in validation set is less than 30.7 seconds, we adopt 768 frames as input and set the input resolution to 112$\times$112 due to the limited GPU memory.

\textbf{Training Setting}
For each input video, we randomly crop the video to generate a clip with 768 frames and at least 75\% of an action instance is guaranteed to be left. 

For image level data augmentations, we set the output resolution of random crop to 112$\times$112. Photo distortion keeps the same as \cite{liu2016ssd}. The angle of random rotation ranges from -45 to +45 and horizontal direction is applied in random flip.

The weights of basic feature extractor are initialized with pre-trained ResNet50 I3D on Kinetics \cite{kay2017kinetics} and other layers are initialized from scratch. We fix the weights of the first two stages and freeze all batch normalization layers of feature extractor for training efficiency. 

We train the model using Stochastic Gradient Descent (SGD) with 0.9 momentum and 0.0001 weight decay. The batch size is set to 16. The schedule of learning rate is annealing down from 0.01 to 0.0001 every 100 epochs out of 1200 epochs using the cosine decay rule. In the first 500 iterations, learning rate linearly warms up from 0.001 to 0.01.

\textbf{Test Setting}
For the inference in test, when processing a long untrimmed video, we first use the sliding window (768 window size) to generate video clips with overlap ratio of 25$\%$ in order, then image center crop is adopted to crop each frame to 112$\times$112 resolution. After DaoTAD generates action segments candidates for all video clips, the candidates with classification scores smaller than 0.005 are filtered. Then NMW \cite{ning2017inception} with threshold 0.5 is applied for each action class separately and the outputs of all classes are merged as the final results.

\subsection{Comparison with the State-of-the-Arts}
We compare our proposed DaoTAD with state-of-the-art methods on THUMOS14 \cite{jiang2014thumos} and summarize the results in Table \ref{tab:compare}. It can be seen that our DaoTAD outperforms all previous one-stage methods. Particularly, DaoTAD achieves improvement of 6.6\% (from 37\% to 43.6\%) on mAP@0.6 and 9\% (from 21.1\% to 30.1\%) on mAP@0.7 compared with MLTPN \cite{wang2020multi}, which is the strongest one-stage competitor on THUMOS14 \cite{jiang2014thumos}. Moreover, our method is comparable with the state-of-the-art two-stage detector AFSD \cite{lin2021learning}. It is worth noting that DaoTAD is based on single RGB stream while MLTPN \cite{wang2020multi} and ASFD \cite{lin2021learning} are based on two-stream input. This fully proves the superiority of our proposed method.

\subsection{Comparison on Inference Time}
As mentioned above, the proposed DaoTAD achieves not only high accuracy, but also high efficiency. Here we benchmark our model on a Nvidia Geforce GTX 1080Ti GPU and present the inference speed measured by fps among other state-of-the-art methods. As shown in Table \ref{tab:fps}, DaoTAD is able to operate at a speed of 6668 fps. In other words, a 60 minutes video with 30 fps can be processed in around 16 seconds. For comparison, existing methods in Table \ref{tab:fps} have significantly lower fps for overall temporal action detection. Notably, these methods are actually much slower in real applications when taking the time cost of optical flow extraction into account. The high efficiency of our DaoTAD benefits from two main facts. First, our model is a one-stage detector and thus eliminates the additional proposal generation stage. Second, our model is based on single RGB stream and thus avoid time-consuming two-stream architecture and optical flow extraction. 

\begin{table*}[htbp]
    \caption{Study of different image level augmentation settings on THUMOS14 in terms of mAP(\%)@tIoU}
    \centering
    \begin{tabular}{ c c c c c | c c c c }
        \toprule
        Baseline & Random Crop & Photo Distortion & Random Rotation & Random Flip & 0.5 & 0.6 & 0.7 & Avg \\
        \midrule
        \ding{51} & - & - & - & - & 48.4 & 38.7 & 24.3 & 37.1 \\
        \ding{51} & \ding{51} & - & - & - & 51.4 & 41.7 & 28.1 & 40.4 \\
        \ding{51} & \ding{51} & \ding{51} & - & - & 52.3 & 42.7 & 29 & 41.3 \\
        \ding{51} & \ding{51} & \ding{51} & \ding{51} & - & 53.3 & 42.7 & 30.1 & 42.0 \\
        \ding{51} & \ding{51} & \ding{51} & \ding{51} & \ding{51} & \textbf{53.8} & \textbf{43.6} & \textbf{30.1} & \textbf{42.5} \\
        \bottomrule
    \end{tabular}
    \label{tab:augmentation}
\end{table*}

\begin{figure*}[htp]
    \centering
    \subfigure[]{
    \label{anchor3}
    \includegraphics[width=8.5cm]{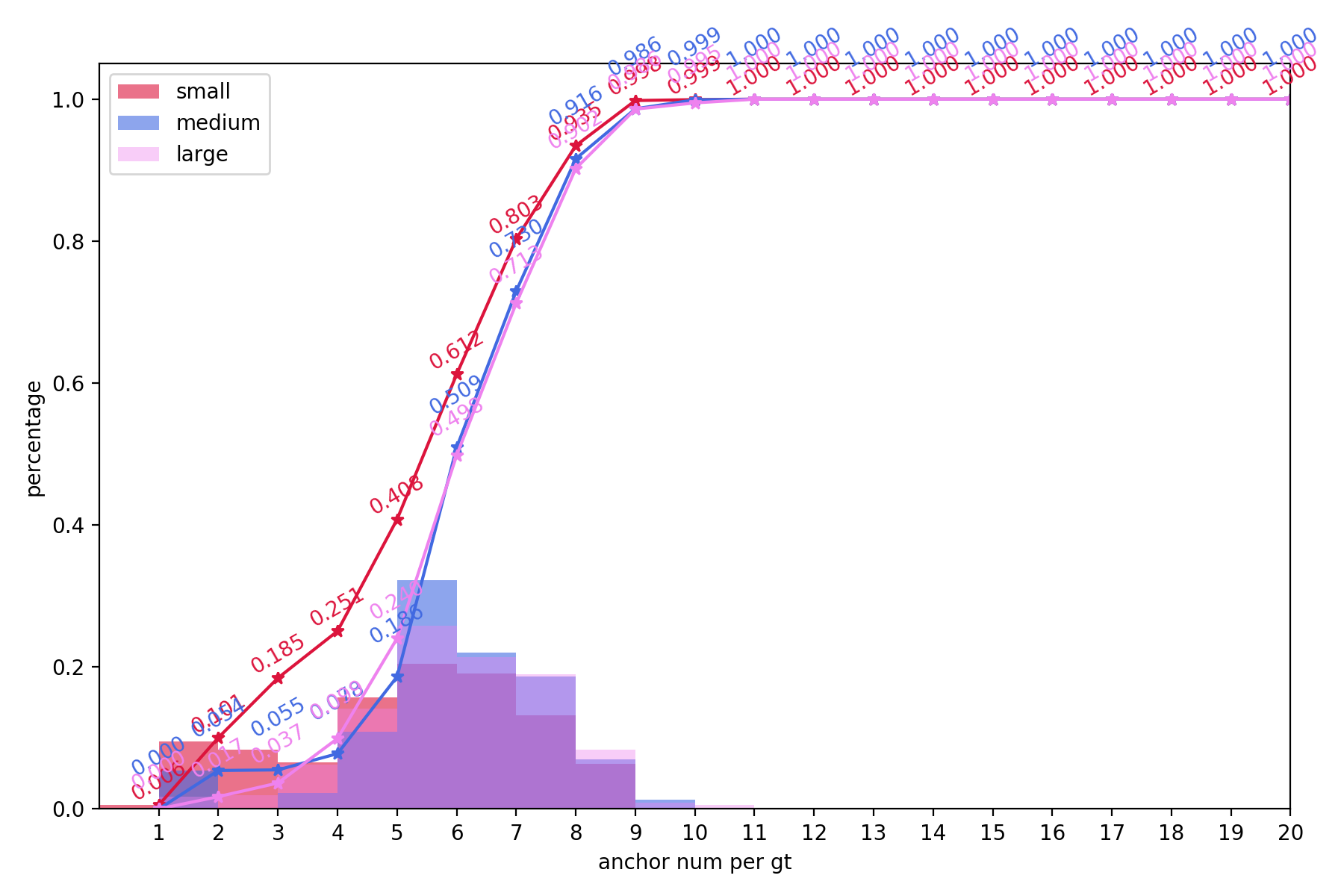}}
    \subfigure[]{
    \label{anchor5}
    \includegraphics[width=8.5cm]{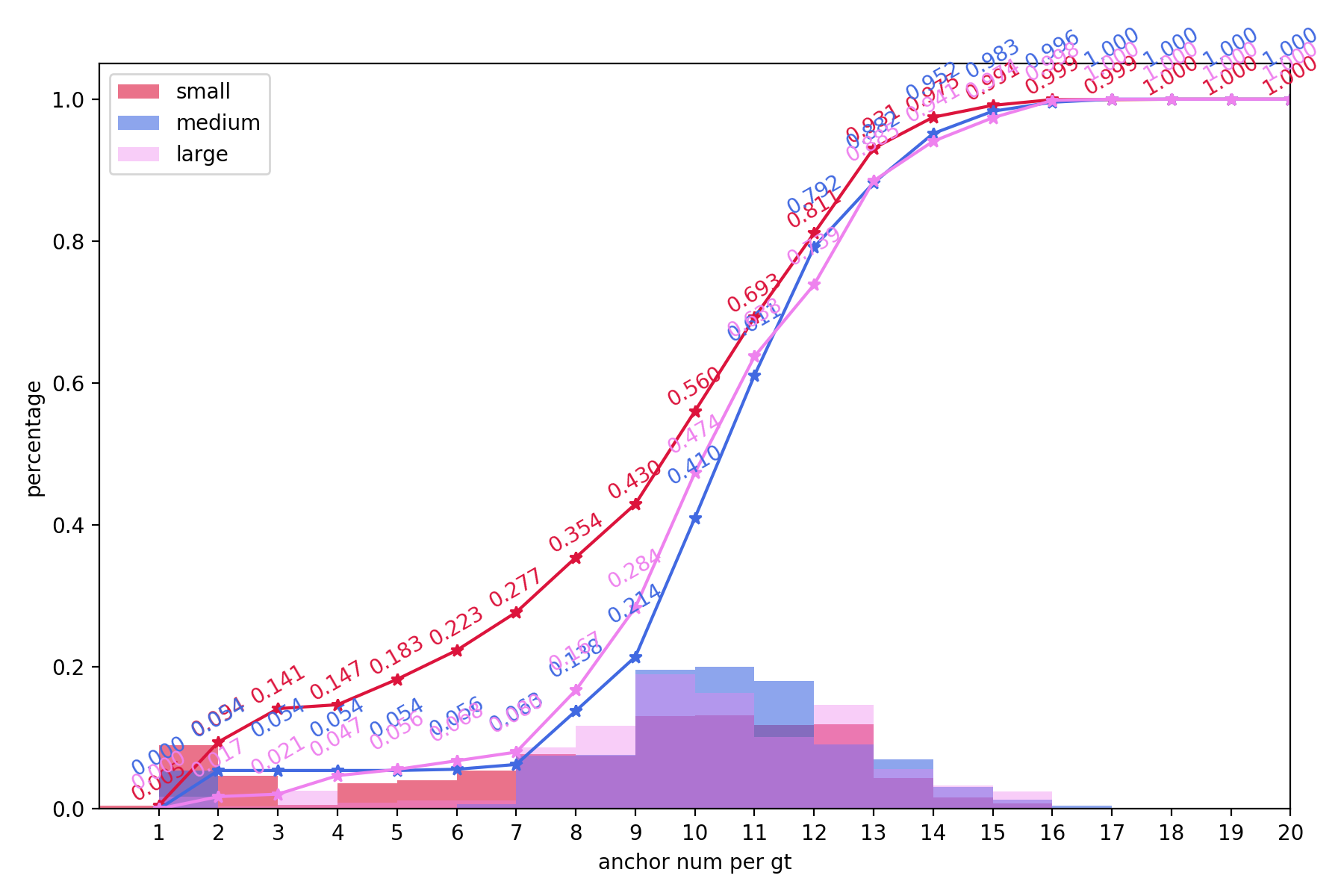}}
    \caption{Cumulative distribution function and probability density function of the number of positive samples assigned to each ground-truth. Different colors represent different scales of ground-truth, we set the scale as $small \in (0s, 2.5s], medium \in (2.5s, 6s], large \in (6s, \infty)$ for THUMOS14 (a) The distribution of 3 anchor setting. (b) The distribution of 5 anchor setting.}
    \label{fig:anchor}
\end{figure*}

\begin{table}[htbp]
    \caption{Study of different anchor settings on THUMOS14 in terms of mAP(\%)@tIoU}
    \centering
    \begin{tabular}{ c | c c c c c c }
        \toprule
        Anchors & 0.5 & 0.6 & 0.7 & Avg \\
        \midrule
        3 & 52.7 & 42.8 & \textbf{30.2} & 41.9 \\
        5 & \textbf{53.8} & \textbf{43.6} & 30.1 & \textbf{42.5} \\
        7 & 53.5 & 42.6 & 29.8 & 42.0 \\
        \bottomrule
    \end{tabular}
    \label{tab:anchor}
\end{table}

\subsection{Ablation Study}
In this section, we conduct several ablation studies on THUMOS14 \cite{jiang2014thumos} to further investigate the efficacy of key components and hyper-parameter settings in our proposed DaoTAD. For all experiments, we only change the corresponding part and use the same evaluation setting.

\textbf{Image Level Data Augmentation}
We compare different image level data augmentation settings and the results are presented in Table \ref{tab:augmentation}. It is obvious that image level data augmentations are crucial for temporal action detection. Without image level data augmentations, the average mAP degrades seriously from 42.5\% to 37.1\%. Moreover, random crop significantly boosts the performance by 3.3\% compared with other data augmentations. We also find these four image level data augmentations are highly complementary.

\textbf{Anchor Setting}
We try to use different number of anchors in our experiments and the results are summarized in Table \ref{tab:anchor}. We can see that using 5 anchors achieves the best average mAP and using more anchors does not obtain better performance due to more redundant candidates being involved. We also conduct the experiment with the default anchor setting (3 anchors) in RetinaNet \cite{lin2017focal} and the average mAP is lower than our anchor setting. To further illustrate the advantage of our anchor setting compared with the one in RetinaNet \cite{lin2017focal}, we utilize the detection analysis tool \footnote{\url{https://github.com/Media-Smart/volkscv}} to get the distributions of positive samples assigned to each ground-truth for the two different anchor settings. As shown in Figure \ref{fig:anchor}, our setting can effectively relieve the imbalance problem as the number of positive assigned samples is more similar across scales. We can conclude that our anchor setting is better than the RetinaNet anchor setting for temporal action detection.

\begin{table}[htbp]
    \caption{Study of different number of frozen stages of backbone on THUMOS14 in terms of mAP(\%)@tIoU}
    \centering
    \begin{tabular}{ c | c c c c }
        \toprule
        Frozen Stages & 0.5 & 0.6 & 0.7 & Avg \\
        \midrule
        4 & 51.4 & 41.2 & 27.3 & 40.0 \\
        3 & 52.6 & 42.7 & 29.6 & 41.6 \\
        2 & \textbf{53.8} & \textbf{43.6} & \textbf{30.1} & \textbf{42.5} \\
        1 & 52.4 & 43.1 & 29.7 & 41.7 \\
        \bottomrule
    \end{tabular}
    \label{tab:frozen}
\end{table}

\begin{table}[htbp]
    \caption{Study of different reduction modes in Spatial Reduction Module on THUMOS14 in terms of mAP(\%)@tIoU}
    \centering
    \begin{tabular}{ c | c c c c }
        \toprule
        Mode & 0.5 & 0.6 & 0.7 & Avg \\
        \midrule
        conv & 52.9 & 43.6 & \textbf{30.2} & 42.2 \\
        max-pool & 53.0 & 42.5 & 29.1 & 41.5\\
        avg-pool & \textbf{53.8} & \textbf{43.6} & 30.1 & \textbf{42.5} \\
        \bottomrule
    \end{tabular}
    \label{tab:SRM}
\end{table}

\begin{table}[htbp]
    \caption{Study of different channel settings in Temporal Downsample Module on THUMOS14 in terms of mAP(\%)@tIoU}
    \centering
    \begin{tabular}{ c | c c c c }
        \toprule
        Channels & 0.5 & 0.6 & 0.7 & Avg \\
        \midrule
        256 & 53 & 42.5 & \textbf{30.2} & 41.9 \\
        512 & \textbf{53.8} & \textbf{43.6} & 30.1 & \textbf{42.5} \\
        1024 & 52.4 & 42 & 28.6 & 41 \\
        \bottomrule
    \end{tabular}
    \label{tab:channel}
\end{table}

\textbf{Frozen Stages}
Freezing more stages of backbone network during training could prevent overfitting while limit the adaptability to new data distribution. Here, we study the effect of different number of frozen stages on the performance. The results in Table \ref{tab:frozen} demonstrate that the average mAP increases 2.5\% when the number of frozen stages decreases from 4 to 2. Further reducing the number of frozen stages has a negative impact on the performance. This indicates that our model benefits from the low level features extracted by early stages with pre-trained weights.

\textbf{Spatial Reduction Module}
In SRM introduced in Section \ref{subsection:SRM}, we adopt the average pooling reduction mode. We compare this one with another two different reduction modes as follow: (1) \textbf{Convolution} A convolutional layer of $1\times4\times4$ kernel without padding is applied to squeeze the spatial dimension of the input 3D feature (2) \textbf{Maximum pooling} The average operation is replaced with the maximum. The results are summarized in Table \ref{tab:SRM}. Among all reduction modes, SRM with average pooling mode achieves the best performance, showing 0.3\%, 1.0\% advantage of average mAP against convolution, maximum pooling respectively.

\textbf{Temporal Downsample Module}
TDM is proposed to generate feature pyramids to handle various temporal scales of action instances. In Table \ref{tab:channel}, we assess the impact of different channel choices of TDM on performance. As can be seen, increasing channels from 256 to 512 has an improvement of the average mAP (from 41.9\% to 42.5\%), while the performance drops from 42.5\% to 41\% after continuing to increase the channels to 1024. This shows that too few channels prevent TDM to extract discriminative features while too many channels cause a mass of parameters which may result in overfitting.

\section{Conclusion}
In the work, we point out that optical flow, a hand-designed representation which not only requires heavy computation, but also makes it methodologically unsatisfactory that two-stream methods are often not learned end-to-end jointly with the flow, is dispensable in high-accuracy temporal action detectors. We discover that image level data augmentation is very important to prevent performance degradation in single RGB stream detector which is neglected in the community. We design an efficient one-stage temporal action detector based on RGB stream named DaoTAD as simple as RetinaNet in object detection. When equipped with image level data augmentation, DaoTAD achieves comparable accuracy to best-performing two-stream temporal action detectors at more than twice the inference speed. Code is available at \url{https://github.com/Media-Smart/vedatad}.
{\small
\bibliographystyle{ieee_fullname}
\bibliography{egbib}
}

\end{document}